\pdfoutput=1

\documentclass[11pt]{article}

\usepackage[]{acl}

\usepackage{times}
\usepackage{latexsym}
\usepackage{graphicx}
\usepackage{amsmath,amsthm,bm}
\usepackage{amsfonts}
\usepackage{booktabs}
\usepackage{multirow}
\usepackage{booktabs}
\usepackage{color}
\usepackage{subfigure}
\usepackage{CJK}

\usepackage[T1]{fontenc}
\usepackage[utf8]{inputenc}

\usepackage{microtype}

\title{Retrieval Oriented Masking Pre-training \\ Language Model for Dense Passage Retrieval}
\author{Dingkun Long, Yanzhao Zhang, Guangwei Xu, Pengjun Xie \\
  Alibaba Group \\
  \texttt{dingkun.ldk,zhangyanzhao.zyz@alibaba-inc.com} \\
  \texttt{kunka.xgw,pengjun.xpj@alibaba-inc.com} \\
}  

\begin{document}
\begin{CJK}{UTF8}{gbsn}
\maketitle
\begin{abstract}
Pre-trained language model (PTM) has been shown to yield powerful text representations for dense passage retrieval task. The Masked Language Modeling (MLM) is a major sub-task of the pre-training process. However, we found that the conventional random masking strategy tend to select a large number of tokens that have limited effect on the passage retrieval task (e,g. stop-words and punctuation). By noticing the term importance weight can provide valuable information for passage retrieval, we hereby propose alternative retrieval oriented masking (dubbed as ROM) strategy where more important tokens will have a higher probability of being masked out, to capture this straightforward yet essential information to facilitate the language model pre-training process. Notably, the proposed new token masking method will not change the architecture and learning objective of original PTM. Our experiments verify that the proposed ROM enables term importance information to help language model pre-training thus achieving better performance on multiple passage retrieval benchmarks.

\end{abstract}

\section{Introduction}
Dense passage retrieval has drown much attention recently due to its benefits to a wide range of down-streaming applications, such as open-domain question answering~\cite{karpukhin-etal-2020-dense,Qu2021RocketQAAO,Zhu2021AdaptiveIS}, conversational systems~\cite{Yu2021FewShotCD} and web search~\cite{Lin2021PretrainedTF,Fan2021PretrainingMI,Long2022MultiCPRAM}. To balance efficiency and effectiveness, existing dense passage retrieval methods usually leverage a dual-encoder architecture. Specifically, query and passage are encoded into continuous vector representations by language models (LMs) respectively, then, a score function is applied to estimate the semantic similarity between the query-passage pair.

Based on the dual-encoder architecture, various optimization methods have been proposed recently, including hard negative training examples mining~\cite{Xiong2021ApproximateNN}, optimized PTMs specially designed for dense retrieval~\cite{Gao2021CondenserAP,Gao2022UnsupervisedCA,Ma2022PretrainAD} and  alternative text representation methods or fine-tuning strategies~\cite{karpukhin-etal-2020-dense,Zhang2022MultiViewDR,DBLP:journals/corr/abs-2110-03611}. In this paper, we focus on studying the part of pre-trained language model. We observe that the widely adopted random token masking MLM pre-training objective is sub-optimal for dense passage retrieval task. Referring to previous studies, introducing the weight of each term (or token) to assist in estimating the query-passage relevance is effective in both passage retrieval and ranking stages~\cite{Dai2020ContextAwareTW,Ma2021BPROPBP,cotmae}. However, the random masking strategy does not distinguish the term importance of tokens. Further, we find that about $40\%$ of the masked tokens produced by the $15\%$ random masking method are stop-words or punctuation\footnote{We used nltk and gensim stop-words lists.}. Nonetheless, the effect of these tokens on passage retrieval is extremely limited~\cite{Fawcett2020LongTermWM}. Therefore, we infer that LMs pre-trained with random token masking MLM objective is sub-optimal for dense passage retrieval due to its shortcoming in distinguishing token importance.   

To address the limitation above, we propose alternative retrieval oriented masking (ROM) strategy aiming to mask tokens that are required for passage retrieval. Specifically, in the pre-training process of LM, the probability of each token being masked is not random, but is superimposed by the important weight of the token corresponded. Here, the important weight is represented as a float number between $0$ and $1$. In this way, we can greatly improve the probability of higher-weight tokens being masked out. Therefore, the pre-trained language model will pay more attention to higher-weight words thus making it more proper for down-streaming dense passage retrieval applications.

To verify the effectiveness and robustness of our proposed retrieval oriented masking method, we conduct experiments on two commonly used passage retrieval benchmarks: the MS MARCO passage ranking and Neural Questions (NQ) datasets. Empirically experiment results demonstrate that our method can remarkably improve the passage retrieval performance. 

\section{Related Work}
Existing dense passage retrieval methods usually adopts a dual-encoder architecture. In DPR~\cite{karpukhin-etal-2020-dense}, they firstly presented that the passage retrieval performance of dense dual-encoder framework can remarkable outperform traditional term match based method like BM25. Based on the dual-encoder framework, studies explore to various strategies to enhance dense retrieval models, including mining hard negatives in fine-tuning stage~\cite{Xiong2021ApproximateNN,Zhan2021OptimizingDR}, knowledge distillation from more powerful cross-encoder model~\cite{Ren2021RocketQAv2AJ,DBLP:journals/corr/abs-2110-03611,ERNIE-Search}, data augmentation~\cite{Qu2021RocketQAAO} and tailored PTMs~\cite{Chang2020PretrainingTF,Gao2021CondenserAP,Gao2022UnsupervisedCA,Ma2022PretrainAD,liu2022retromae,cotmae}. 

For the pre-training of language model, previous research attend to design additional pre-training objectives tailored for dense passage retrieval~\cite{lee2019latent,Chang2020PretrainingTF} or adjust the Transformer encoder architecture~\cite{Gao2021CondenserAP,Gao2022UnsupervisedCA} to obtain more practicable language models. In this paper, we seek to make simple transformations of the original MLM learning objective to improve the model performance, thereby reducing the complexity of the pre-training process.

\begin{figure*}
    \centering
    \includegraphics[width=2.0\columnwidth]{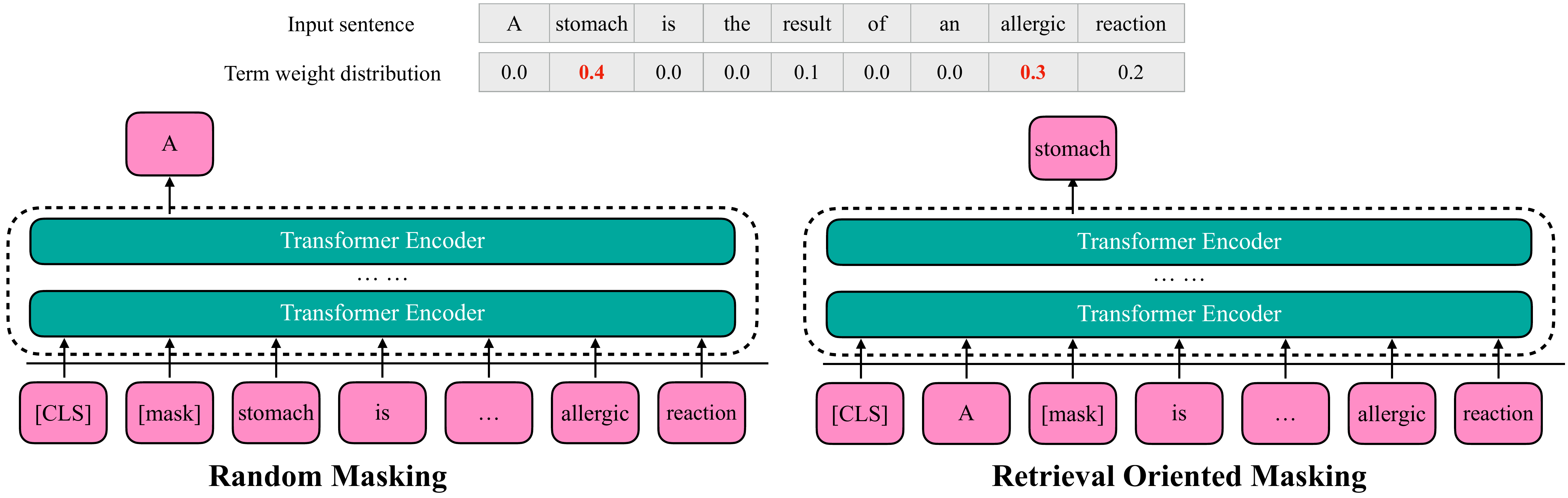}
    \caption{An illustration of our retrieval oriented masking (ROM) method. ``stomach'' is the token with the highest term weight in the input sentence, thus sharing a larger probability of being masked out.}
    \label{fig:rom-model}
\end{figure*}

\section{Methodology}
In this section, we describe our proposed pre-training method for the dense passage retrieval task. We first give a brief overview of the conventional BERT pre-training model with MLM loss. Then we will introduce how to extend it to our model with retrieval oriented masking pre-training. 

\subsection{BERT Pre-trained Model}

\noindent{\textbf{MLM Pre-training}} Many popular Transformer Encoder language models (e,g. BERT, RoBERTa) adopts the MLM objective at pre-training phase. MLM masks out a subset of input tokens and requires the model to predict them. Specifically, the MLM loss can be formulated as:
\begin{equation*}
    \mathcal{L}_{mlm} = \sum_{i\in masked} {\rm CrossEntropy}(Wh_i^{L}, x_i),
\end{equation*}
where $h_i^{L}$ is the final representation of masked token $x_i$ and $L$ is the number of Transformer layers.

\vspace{0.2cm}
\noindent{\textbf{Random Masking}} In general, the selection of masked out tokens is random, and the proportion of masking in a sentence is set at $15\%$. Mathematically, for each token $x_i \in \boldsymbol x$, the probability of $x_i$ being masked out $p(x_i)$ is sampled from a uniform distribution between $0$ and $1$. If the value of $p(x_i)$ is in the top $15\%$ of the entire input sequence, then $x_i$ will be masked out.

\subsection{Disadvantages of Random Masking} 
The significant issue of the random masking method is that it does not distinguish the important weight of each token. Statistic analysis illustrates that $40\%$ of the tokens masked by the random masking strategy are stop-words or punctuation. As shown in previous studies, it is valuable to distinguish the weights of different terms for passage retrieval. Whether for the query or passage, terms with higher important weights should contribute more to the query-passage relevance estimation process. Although the pre-train language model itself is contextualized aware, we still hope that the language model has a stronger feature of distinguishing term importance for retrieval task. However, the language model trained by the random masking strategy is flawed. 


\subsection{Retrieval Oriented Masking} 
As mentioned above, term importance is instructive for passage retrieval. Here, we explore to introduce term importance into the MLM training. More specifically, we incorporate the term importance information into token masking. Different from the random masking strategy, whether a token $x_i$ is masked is not only determined by the random probability $p_r(x_i)$, but also determined by its term weight $p_w(x_i)$. Here, $p_w(x_i)$ is normalized between value $0$ and $1$. The final probability of token $x_i$ being masked out is $p_r(x_i)$ + $p_w(x_i)$.

Then the problem now is to calculate the term weight of each token. Previous studies have proposed different methods to calculate word weights~\cite{Mallia2021LearningPI,Ma2021BPROPBP}, which can be roughly divided into unsupervised and supervised categories. To maintain the unsupervised pre-training paradigm of LM, we adopt the unsupervised method presented in the BPROP work~\cite{Ma2021BPROPBP}. 

The BPROP proposed to compute the term weight distribution in a sentence based on BERT's vanilla [CLS]-Token attention weights by considering that the token [CLS] is an aggregate of the entire sequence representation. However, the term distribution obtained from BERT's vanilla attention is a semantic distribution, but not an informative distribution. BPROP leverages a contrastive method to produce the final distribution. Formally, Given an input sentence $\boldsymbol x = (x_1, x_2, \cdots , x_n )$, let $a_i$ denotes the attention weight of $x_i \in x$ for the [CLS] token which is calculated as an average of each head's attention weights from the last layer of BERT. The BPROP method will produce a new contrastive term distribution $p_w(x)$ in an totally unsupervised manner based on $(a_1, a_2, \cdots , a_n)$, where $\sum_i^n p_w(x_i|\theta_{\rm BPROP})$ $=1$ . Here, we omit the specific calculation process and more details of BPROP can be found in the original paper~\cite{Ma2021BPROPBP}.

Once we calculate the term weight distribution of each sentence in the corpus in advance, we can conduct the LM pre-training with MLM learning objective by our ROM strategy. It should be noted that in the ROM method, the masked probability of each token still relies on the uniform random probability since we still want to keep the basic properties of the pre-trained LM, rather than let the LM only focus on a small number of higher-weight tokens. In practice, the proportion of mask tokens is also set to $15\%$, and statistical analysis show that the proportion of stop-words and punctuation token masked out in the ROM method dropped to $14\%$.

\section{Experiments}
\subsection{Datasets} We evaluate the proposed model on the following data sets. \textbf{MS MARCO Passage Ranking} is a widely used benchmark dataset for passage retrieval task, and which is constructed from Bing's search query logs and web documents retrieved by Bing~\cite{Campos2016MSMA}. \textbf{Neural Question} is another passage retrieval dataset derived from Google search~\cite{Kwiatkowski2019NaturalQA}. For each dataset, we follow the standard data splits in previous work~\cite{Gao2022UnsupervisedCA}.

\subsection{Compared Methods} To verify the effectiveness of our proposed method, we adopt the following methods which focused on PTM optimization as our main baselines: \textbf{Condenser}~\cite{Gao2021CondenserAP} adapts the Transformer architecture in LM pre-training to enhance the text representations thus facilitating down-streaming passage retrieval; \textbf{coCondenser}~\cite{Gao2022UnsupervisedCA} is an extension of Condenser, which uses an unsupervised corpus-level contrastive loss to warm up the passage embedding space; \textbf{COSTA}~\cite{Ma2022PretrainAD} introduces a novel contrastive span prediction task in LM pre-training aiming to build a more discriminative text encoder.

We directly borrowed several other competitive baselines from the coCondenser paper, including lexical systems BM25, DeepCT~\cite{Dai2020ContextAwareTW}, DocT5Query~\cite{Cheriton2019FromDT} and GAR~\cite{Mao2021GenerationAugmentedRF}; and dense systems DPR~\cite{karpukhin-etal-2020-dense}, ANCE~\cite{Xiong2021ApproximateNN}, and ME-BERT~\cite{Luan2021SparseDA}.

\begin{table*}
\caption{Experiment Results for MS MARCO Passage Retrieval and Natural Question Datasets. T-test demonstrates the improvements of ROM and coROM to the baselines are statistically significant ($p\leq 0.05$).}
\label{tab:my-table}
\centering\centering
\begin{tabular}{c|cc|ccc}
\toprule
\multirow{2}{*}{Method}          & \multicolumn{2}{c|}{MS MARCO Passage}                   & \multicolumn{3}{c}{Neural Question}                                   \\ \cmidrule(l){2-6} 
                                 & MRR@10                   & \multicolumn{1}{c|}{R@1000} & R@5                   & R@20                  & R@100                 \\ \midrule
BM25        & 18.6                        & \multicolumn{1}{c|}{85.7}   & -                     & \multicolumn{1}{c}{59.1}                  & \multicolumn{1}{c}{73.7}                  \\
\multicolumn{1}{c|}{DeepCT~\cite{dai2019context}}      & \multicolumn{1}{c}{24.3}                     & \multicolumn{1}{c|}{-}      & -                     & -                     & -                     \\
\multicolumn{1}{c|}{DocT5Query~\cite{Cheriton2019FromDT}}  & \multicolumn{1}{c}{27.7}                     & \multicolumn{1}{c|}{94.7}   & -                     & -                     & -                     \\
\multicolumn{1}{c|}{GAR~\cite{Mao2021GenerationAugmentedRF}}         & -                        & \multicolumn{1}{c|}{-}      & \multicolumn{1}{c}{60.9}                  & \multicolumn{1}{c}{74.4}                  & \multicolumn{1}{c}{85.3}                  \\ \midrule
\multicolumn{1}{c|}{DPR~\cite{karpukhin-etal-2020-dense}}         & -                        & \multicolumn{1}{c|}{-}      & -                     & \multicolumn{1}{c}{74.4}                  & \multicolumn{1}{c}{85.3}                  \\
\multicolumn{1}{c|}{BERT$_{base}$}         & 33.4                       & \multicolumn{1}{c|}{95.5}      & -   & \multicolumn{1}{c}{-}  & \multicolumn{1}{c}{-}                  \\
\multicolumn{1}{c|}{ANCE~\cite{Xiong2021ApproximateNN}}        & \multicolumn{1}{c}{33.0}                     & \multicolumn{1}{c|}{95.5}   & -                     & \multicolumn{1}{c}{81.9}                  & \multicolumn{1}{c}{87.5}                  \\
\multicolumn{1}{c|}{ME-BERT~\cite{Luan2021SparseDA}}     & \multicolumn{1}{c}{33.8}                     & \multicolumn{1}{c|}{-}      & -                     & -                     & -                     \\
\multicolumn{1}{c|}{RocketQA~\cite{Qu2021RocketQAAO}}    & \multicolumn{1}{c}{37.0}                     & \multicolumn{1}{c|}{97.9}   & \multicolumn{1}{c}{74.0}                  & \multicolumn{1}{c}{82.7}                  & \multicolumn{1}{c}{88.5}                  \\
\multicolumn{1}{c|}{Condenser~\cite{Gao2021CondenserAP}}   & \multicolumn{1}{c}{36.6}                     & \multicolumn{1}{c|}{97.4}   & -                     & \multicolumn{1}{c}{83.2}                  & \multicolumn{1}{c}{88.4}                  \\
\multicolumn{1}{c|}{COSTA~\cite{Ma2022PretrainAD}}       & \multicolumn{1}{c}{36.6}                    & \multicolumn{1}{c|}{97.1}  & -                     & -                     & -                     \\ 
\multicolumn{1}{c|}{\bf{ROM}}         & \multicolumn{1}{c}{37.3} & \multicolumn{1}{c|}{98.1}      & \multicolumn{1}{c}{73.9} & \multicolumn{1}{c}{ 83.4} & \multicolumn{1}{c}{88.5} \\ \midrule
\multicolumn{1}{c|}{coCondenser~\cite{Gao2022UnsupervisedCA}} & \multicolumn{1}{c}{38.2}                     & \multicolumn{1}{c|}{98.4}   & \multicolumn{1}{c}{75.8}                  & \multicolumn{1}{c}{84.3}                  & \multicolumn{1}{c}{\bf 89.0}                  \\ 
\multicolumn{1}{c|}{\bf{coROM}}        & \multicolumn{1}{c}{{\bf 39.1}} & \multicolumn{1}{c|}{{\bf 98.6}}      & \multicolumn{1}{c}{{\bf 76.2}} & \multicolumn{1}{c}{{\bf 84.6}} & \multicolumn{1}{c}{88.8} \\ \bottomrule
\end{tabular}
\end{table*}

\subsection{Experimental Setup}
Our ROM language model pre-training starts with a vanilla 12-layer BERT-base model. Following previous work, we use the same data as BERT in pre-training: English Wikipedia and the BookCorpus. In addition, like the coCondenser model, we also train a language model that adds a contrastive learning loss function based on the ROM model and target corpus (Wikipedia or MS-MARCO web collection). Here, the co-training drove ROM model is denoted as coROM\footnote{The fine-tuned model on MS MARCO passage ranking dataset is available at \url{https://modelscope.cn/models/damo/nlp_corom_sentence-embedding_english-base/summary}. The original ROM and coROM models will be publicly available in the future.}.

During the fine-tuning process, we adopt the AdamW optimizer with $5e$-$6$ learning rate and batch size $64$ for $3$ epochs for the MS MARCO passage dataset. For the NQ dataset, we follow the hyper-parameter setting presented in the DPR toolkit~\cite{karpukhin-etal-2020-dense}. For the MS MARCO passage dataset, the test set label is not available, we only report results on the dev set. We follow the evaluation methodology used in previous work~\cite{Gao2022UnsupervisedCA}. All experiments are conducted on 8 NVIDIA Tesla 32G V100.

\subsection{Evaluation Results}
The overall performance of all baselines and ROM are reported in Table \ref{tab:my-table}. The results indicate that ROM achieves the best performance. 
Firstly, the improvement of the ROM model is extremely significant compared with the vanilla BERT model. For example, the MRR@10 metric on the MS MARCO dataset has increased from $33.4$ to $37.3$, which empirically proves the benefits of tailored LM. 
Compared with other dense retrieval tailored LMs, the ROM model achieves consistent improvement on both two datasets. 
Additionally, similar to the coCondenser model, coROM model will further improve the passage retrieval performance with the help of the contrastive co-training method, indicating that high-quality text representation is the foundation of dense passage retrieval.

\begin{table}[h]
\centering
\caption{MRR@10 metric on the MS MARCO Passage Ranking leaderboard. We bold the best performances of both the dev and eval set.}
\label{tab:learderboard}
\resizebox{1.0\columnwidth}{!}{
\begin{tabular}{@{}c|cc@{}}
\toprule
\multirow{1}{*}{Model} & dev & eval       \\ \midrule
Search LM (SLM) + HLATR  & {\bf 46.3}  & {\bf 45.0}          \\
Listwise + Fusion reranker	& 45.4 & 44.0  \\
Cot-MAE~\cite{cotmae} & 45.6 & 43.8 \\
Lichee-xxlarge + deberta-v3-large & 45.2 & 43.6  \\
Adaptive Batch Scheduling~\cite{9511263}	 & 45.3 & 43.1  \\
coCondenser~\cite{Gao2022UnsupervisedCA} & 44.3 & 42.8 \\ \bottomrule
\end{tabular}}
\end{table}

\subsection{MS MARCO Passage Ranking LeaderBoard}

To further verify the effectiveness of the ROM family model, we conduct an experiment with full retrieval-then-reranking pipeline and submit our result to MS MARCO LeaderBoard. Table~\ref{tab:learderboard} presents the top systems on the MS MARCO Passage Ranking leaderboard\footnote{\url{https://microsoft.github.io/msmarco/}}. For the model description ``Search LM (SLM) + HLATR'', ``Search LM(SLM)'' is the coROM model and HLATR~\cite{zhang2022hlatr} is a lightweight reranking module coupling both retrieval and reranking features thus further improving the final ranking performance. The final submission is an ensemble of multiple ``reranking + HLATR'' models trained on different pretrained language models (e,g. BERT, ERINE and RoBERTa).  


\subsection{Analysis}
\noindent\textbf{Quality of Term Weights} Intuitively, the quality of term weights will directly affect the performance of the ROM model and the supervised method can produce higher quality term weight results. Thus, in addition to the BPROP method, we also conduct ROM pre-training by term weight distribution generated by the supervised DeepImpact method~\cite{Mallia2021LearningPI}. The experimental results of MS MARCO passage dataset are presented in Table \ref{tab:rom-term-weights}. From which we can observe that: 1) High-quality term weight results do lead to better passage retrieval performance; 2) The improvement brought by DeepImpact is much smaller than that of the ROM model compared to the vanilla BERT LM, which indicates that the unsupervised term weight computation method is decent by considering that supervised method will inevitably introduce extra training cost.

\begin{table}[]
\centering\centering
\small
\caption{ROM results on MS MARCO passage dataset with different methods for producing term weights.}
\label{tab:rom-term-weights}
\begin{tabular}{c|c|c}
\hline
Term Wights & MRR@10 & R@1000 \\ \hline
BPROP       & 37.3   & 98.1   \\ \hline
DeepImpact  & 37.6   & 98.2   \\ \hline
\end{tabular}
\end{table}

\vspace{0.1cm}
\noindent\textbf{Attention Weights Analysis} To verify that the proposed ROM model is more discriminative for tokens with different weights, we compare the [CLS]-token distribution of the ROM and vanilla BERT models. In table \ref{tab:term-weights}, we present the top term weight tokes produced by these two different models. We can observe that the two sets have overlapping tokens. However, the higher attention weight tokens produced by the ROM model is obviously more reasonable, and the token set of the BERT model even contains stop-word marks.

\begin{table}[]
\centering \centering
\small
\caption{Top attention weight tokens produced by BERT and ROM model.}
    \begin{tabular}{p{0.95\columnwidth}}
    \toprule
     Text: A business analyst's daily job duties can vary greatly, depending on the nature of the current organization and project ...\\
     \midrule
     BERT: ., the, analyst, organization, duties \\
     \midrule 
     ROM: business, vary, duties, analyst, depending \\
     \bottomrule
    \end{tabular}
    \label{tab:term-weights}
\end{table}

\section{Conclusion and Future Work}
In this paper, we investigate that current random token masking MLM pre-training method is sub-optimal for LM pre-training, as this process tends to focus on stop words and punctuation. We suggested a novel retrieval oriented masking strategy which incorporates term importance information. We evaluated our ROM and extended coROM LMs on two benchmarks. The results showed that our method is highly effective, and our final model can achieve significant improvements compared to previous tailored LMs for dense passage retrieval. 

In this paper, we intuitively use the BPROP method for term weight computation, and have not compared it with other unsupervised term weight methods. More detailed studies of term weight distribution based on the ROM model may produce an in-depth understanding. In further, we only conduct experiments based on $\rm BERT_{base}$ model, and validation based on extensive LMs pre-trained with MLM objective (e,g. ${\rm RoBERTa}$) can further help to demonstrate the generality of the proposed ROM method. 

\bibliography{rom}
\bibliographystyle{acl_natbib}

\clearpage


\end{CJK}
\end{document}